\documentclass[runningheads,a4paper]{llncs}

\usepackage{todonotes}
\usepackage{graphicx}
\usepackage{booktabs}
\usepackage{tabularx}
\usepackage{wrapfig}
\usepackage[hyphens]{url} 
\usepackage{microtype}

\newcolumntype{Y}{>{\RaggedRight\arraybackslash}X} 
\begin{document}
\title{Adding Why to What? Analyses of an Everyday Explanation} 
%
%
\author{Lutz Terfloth\orcidID{0000-0003-1134-5090} \and
Michael Schaffer\orcidID{0009-0001-5821-9967} \and
Heike M. Buhl\orcidID{0000-0002-1001-492X} \and 
Carsten Schulte \orcidID{0000-0002-3009-4904}}

%
\authorrunning{L. Terfloth et al.}
%
\institute{Paderborn University, Paderborn, Germany\\
\email{\{lutz.terfloth, michael.schaffer, heike.buhl, carsten.schulte\}@uni-paderborn.de}}
\maketitle              

\begin{abstract}
In XAI it is important to consider that, in contrast to explanations for professional audiences, one cannot assume common expertise when explaining for laypeople. But such explanations between humans vary greatly, making it difficult to research commonalities across explanations. We used the dual nature theory, a techno-philosophical approach, to cope with these challenges. According to it, one can explain, for example, an XAI's decision by addressing its dual nature: by focusing on the Architecture (e.g., the logic of its algorithms) or the Relevance (e.g., the severity of a decision, the implications of a recommendation). We investigated 20 game explanations using the theory as an analytical framework. We elaborate how we used the theory to quickly structure and compare explanations of technological artifacts. We supplemented results from analyzing the explanation contents with results from a video recall to explore how explainers  justified their explanation. We found that explainers were focusing on the physical aspects of the game first (Architecture) and only later on aspects of the Relevance. Reasoning in the video recalls indicated that EX regarded the focus on the Architecture as important for structuring the explanation initially by explaining the basic components before focusing on more complex, intangible aspects. Shifting between addressing the two sides was justified by explanation goals, emerging misunderstandings, and the knowledge needs of the explainee. We discovered several commonalities that inspire future research questions which, if further generalizable, provide first ideas for the construction of synthetic explanations.

\keywords{Analysis of Human Explanations \and Naturalistic Explanations \and
Qualitative Analysis \and
Technological Artifacts.}
\end{abstract}

\section{Introduction}
\label{sec:introduction}
Enabling laypeople to “effectively understand, trust, and manage” \cite{gunning_xaiexplainable_2019} AI applications requires XAI systems that provide \textit{understandable explanations}. The General Data Protection Regulation (GDPR) calls for "meaningful explanations of the logic involved" in the context of automated decision-making, too. However, in many instances, experts rely on a notion of what constitutes an explanation for the implementation \cite{van_der_waa_evaluating_2021}. We see potential in adhering to an empirically grounded conceptualization of explanations instead. If an XAI system could mimic how explanations evolve in naturalistic, everyday settings, it could be used to make an AI's output more understandable, especially for laypeople unfamiliar with the technology.

Over the years, various research has been done on explanations \cite{miller_explanation_2019,el-assady_towards_2019}. Although less research focused on explanations of technological artifacts, it is already used in the context of explaining and understanding, yet is still facing a lack of precision \cite{de_ridder_mechanistic_2006}. However, one central idea can be found across different publications in the area: Technological artifacts are human-made objects engineered to serve as means to certain ends, and therefore explaining technological artifacts can be done addressing the two sides of their dual nature \cite{kroes_technological_1998,vermaas_technical_2006}. Whenever, for example, engineers explain their inventions in texts, they are explaining aspects about the physical properties (e.g., the shape, the algorithms, or the logic; its Architecture) of the artifact in alignment with its functional capabilities (e.g, how it serves as a means to an end; its Relevance). In the context of XAI, in which synthetic explanations are constructed by technological artifacts, and based on decisions by technological artifacts, these peculiarities of their dual nature cannot be neglected. We believe, the theory provides a rich background for research on everyday explanations of technological artifacts.

Many scholars agree that knowledge about the dual nature of a technological artifact is relevant for understanding it, yet are undecided whether both sides are equally important or whether one side may be the precursor for the other \cite{soloway_learning_1986,schulte_block_2008}. However, to date, no study investigated specifically if and how the dual nature is addressed in naturalistic everyday explanations of technological artifacts. Thus, how the dual nature is addressed in explanations, which of the two sides are explained first, more frequently, or whether one or the other should be addressed in more detail or omitted completely, is not empirically investigated yet. An empirically grounded conceptualization of how technological artifacts are naturally explained in a verbal, interactive settings would be a useful prerequisite for many research areas, but especially for XAI and education in general. It could provide a foundation for formulating recommendations on whether and which of the artifacts properties should be referred to within a (synthetic) explanation, and thus improve the understandability and interpretability of XAI systems.

This paper is a step towards addressing these shortcomings and investigates how the content of naturalistic explanations evolves and is justified by the explainers (EX). It presents results from a study of dyadic explanatory dialogues in which a technological artifact, the board game \textit{Quarto!}, is explained in a naturalistic setting. The contribution of this paper is twofold. We elaborate on the theoretical foundation used for an empirical study, and present results from that study. It thus consists of two parts: First, in the Sections Background and Study, we provide a brief introduction to the theoretical foundation of the research and connect it to the research questions of the paper. Furthermore, we elaborate how and why we use the dual nature theory as a framework. In the Sections Method, Results and Discussion, the study, data acquisition and analysis are elaborated after which the results are presented and discussed.

\section{Background}
\label{sec:background}
Considerable research attention has been directed towards the conceptualization of explanations from different perspectives and disciplines. A philosophical standpoint sheds light on quality characteristics of explanations and tries to identify factors that contribute to sound and satisfying explanations \cite{salmon_four_1990,chin-parker_contrastive_2017}. Social psychology examines why and when individuals explain \cite{heider_psychology_1958,chin-parker_contrastive_2017,malle_how_2004,gilbert_models_1998}. Explanations can cover a vast array of topics and ideas, but regardless of content diversity, explanations in their composition are shaped by causal and logical factors \cite{salmon_four_1990}, too. Cognitive sciences are investigating important elements of explanations, the process of explanation generation and which cognitive processes succumb to explanatory processes \cite{chin-parker_contrastive_2017,lombrozo_functional_2006}. 

During extemporaneous explanations, EX are facing various challenges to give meaningful, and complete and precise explanations that are well-structured \cite{roscoe_self-monitoring_2014}. Structuring explanations require organization of knowledge \cite{roscoe_self-monitoring_2014}. Therefore, explaining is a cognitive process, that is highly constructive \cite{chi_self-explaining_2000,hale_explanation_1995} and utilizes existing domain knowledge \cite{lombrozo_functional_2006}, that is part of mental representations \cite{chi_self-explaining_2000,keil_explanation_2006}. Thus, EX need to decide on the course of explanations, not only which aspects of a domain they want to explain but especially at what point of time, for example by monitoring the explainee (EE), and mentioned and unmentioned domain aspects \cite{levelt_speakers_1981}. 

These requirements an EX faces makes researching such naturalistic explanations challenging as well as necessary, if the goal is to provide recommendations for XAI systems. To circumvent some of these issues, shared properties of explananda are interesting. The content of an explanation is influenced by the explanandum (the subject of explanation), more precisely the EX's mental representation of that explanandum, and the knowledge needs of the Explainee (EE) regarding that explanandum \cite{chi_can_2004,brennan_partner-specific_2009}. Whenever the explanandum is a technological artifact, one can refer to its dual nature, following a techno-philosophical theory \cite{kroes_technological_1998,vermaas_philosophy_2011}. According to the dual nature theory, an explanation of a technological artifact -- an object engineered to be a means to an end -- can be formulated by addressing two properties \cite{kroes_dual_2002}. On the one hand, a structural mode of description, "makes use of concepts from physical laws and theories and is free of any reference to the function of the object" and addresses the Architecture of the artifact \cite{kroes_technological_1998,schulte_duality_2008}. On the other hand, the functional/teleological mode of description is a way in which “[w]ith regard to its function, a technological object is described in an intentional (teleological) way" \cite{kroes_technological_1998} and addresses the Relevance of the artifact.

Interestingly, in the context of this dual nature theory, the idea of an engineer's ability to "bridge the gap" \cite{kroes_design_2002} between the two sides of the dual nature resulting from their deep understanding of the artifact is discussed. Not only, but especially because  of this, the dual nature theory found application within the computing education community  \cite{schulte_duality_2008},  as well as in the technology education community in the context of understanding technological artifacts \cite{cederqvist_seeing_2021}. It guided a development of an analytical framework to investigate pupil's understanding of programmed technical solutions (PTS), especially concerning their ability to open the black boxes \cite{cederqvist_pupils_2020}. The results indicated that both sides – structure and function – need to be understood to have a more profound understanding of a PTS: "[t]hus, these key elements are important to consider in pedagogical practice to promote learning regarding PTS." \cite{cederqvist_pupils_2020}. The theory served as a theoretical foundation for a program comprehension model, too \cite{schulte_block_2008}. On a more abstract level, comparable ideas were referred to using different foci and terms, but similar ideas in the context of program comprehension: "mechanisms" vs. "explanations" \cite{soloway_learning_1986}, and "text base" vs. "situation model" \cite{pennington_stimulus_1987}. \cite{attisano_components_2021} investigated how children discuss a machine they got to know while visiting a museum, concerning which prompt resulted in the children addressing either the mechanisms or the components when asked questions about the machine. In summary, the main ideas of the dual nature theory have a rich history not only within and outside the computing community, but in the context of making sense of technological artifacts in general.

Whether the dual nature theory is useful for the analysis of explanation content whenever a technological artifact is explained, still needs investigation. We call the sides of the dual nature, the \textit{Architecture} (roughly: what it is, what components are) and \textit{Relevance} (roughly: why and what it and its components are for) of the artifact. Our conceptualization of the two sides is as follows: one can (1) explain how a technological artifact works on the level of data, or the logic of the algorithms (i.e., addressing the Architecture), or by explaining (2) how one may use it for relaxed journey across foreign countries (i.e., addressing the Relevance). But there is uncertainty how the dual nature of the technological artifact influences the structure and evolution of an explanation, and whether these sides can be clearly identified within naturalistic explanations. This paper aims at answering three questions:
\begin{itemize}
    \item (RQ1) How is the dual nature of the explanandum expressed in the utterances in the explanations?
    \item (RQ2) What are common patterns regarding sequences of utterances about the dual nature across the explanations?
    \item (RQ3) How do EX justify their choice of explanation content regarding which side of the dual nature was addressed?
\end{itemize}


\section{Method}
Given our overarching objective to analyze how EX address the sides of the dual nature in their explanations, we conducted a study in which a dyad of people engaged in a naturalistic explanation scenario. To acquire comparable explanations, a study design that allowed to control certain variables that would not be controllable with a more complex explanandum and in the field, was developed. Even though the naturalness of the explanations in the field would be higher, at this early stage of researching the phenomenon a certain uniformity is preferred. Researchers and research assistants adhered to a predefined study guideline. The study was conducted at Paderborn University as well as Bielefeld University in Germany.  This study is part of a larger, more complex study consisting of five different phases: (1) pretesting, (2) explaining, (3) video recall, (4) playing and (5) post testing. This paper focuses on the analysis of phase 2 and 3 (explanation phase and video recall). We recorded the dyads during the explanation. To observe the explanations from another room, we placed a webcam for live-streaming, too. Prior to the explanation, every participant filled out a questionnaire for sociodemographic details. After the explanation (phase 2), we conducted a video recall to assess the EX underlying thinking and reasoning during the explanations based on questions of a structured interview. The analyses of the content of the explanations, and the justifications mentioned in the video recalls, is guided by the dual nature theory as an analytical framework. 
\subsection{Participants and Recruitment} 
\label{sec:sample}
We relied primarily on onsite recruiting via handing out flyers on campus, hanging posters, and hearsay. We compensated all participants monetarily (€10 per hour of participation). Participants signed up voluntarily for the study. All studies included COVID-19 safety measures and obtaining written informed consent before the studies. Communication with participants was based on boilerplate to ensure that all participants had the same information and instructions. We divided a total of 48 participants into two groups: EX and EE (24 dyads). Interaction between EE and EX before the study was prevented as much as possible by meeting and picking them up independently of each other on the day of the study. Within mail communication, we instructed the group of EX to make themselves familiar with the explanandum. They had the option to either pick up a physical copy of the game from our lab or use an online version of the game. EE received only direction, and the location of the study. Exclusion criteria were (1) prior knowledge of the game \textit{Quarto!} (for EE), (2) non-C2 level language skills (for both EE and EX), or (3) prior participation and knowledge about the overarching research endeavor. Four dyads were excluded from the corpus due to a language barrier, two instances in which the EE had prior knowledge of the game, and one instance in which the game was mistakenly present during explanation. The final sample of video data consists of 20 lab studies in which 20 EX explained the game to 20 EE (18 female, 19 male, 1 non-binary)\footnote{3 participants provided no information and are therefore not included in the descriptive statistics.}. Age ranged from 18 to 39 (\textit{M}=24.92, \textit{SD}=4.42). 36 participants had an academic background. 35 reported to be students (e.g., engineering, education, economics, law, computer science, media sciences, linguistics), 2 were full-time employees. Out of 19 EX, 7 reported to have explained the game to someone else before the study. EX reported to have between 0 and 18 rounds of gameplay experience (\textit{M}=5.46, \textit{SD}=5,18). 10 EX reported to have experience in explanations (e.g., tutoring). The studies were planned to take 2 to 3 hours for all assessments. Phase 2 (explanation) varied from 02:23 mm:ss to 16:17 mm:ss (\textit{M}=07:24, \textit{SD}= 03:22). 
\subsection{The Explanandum: \textit{Quarto!} }
\label{sec:quarto}
The explanandum of the study is the strategical, two-player board game \textit{Quarto!}. \textit{Quarto!} is a game developed by Swiss mathematician Blaise Müller. It demands deductive reasoning and thinking, similar to chess. At any moments during the game, all information is openly available to every player. The game is made up of 16 pieces and a board consisting of 4×4 squares. Each of the 16 pieces, that one places on the board, is unique and differs in at least one of the following traits: size (tall or short), shade (light or dark), shape (round or square), and solidity (solid or hollow). The goal of the game is similar to Connect Four: connect a row, column, or diagonal of four pieces that share \textit{at least one trait}. Contrary to the way connect four is played, a player does not choose which piece they place themselves, but instead chooses the piece the opponent has to place in their next move. After the opponent places the piece handed to them, they choose a piece, that the other person has to play. Turn-by-turn, the game continues until either a \textit{Quarto!} is called, and a person wins, or all squares are occupied and the game ends in a draw.

We chose a game and specifically \textit{Quarto!} as the explanandum for several reasons. As our interdisciplinary team strives to triangulate the results from different approaches to get a rich perspective of the phenomenon of natural explanations of technological artifacts, certain tradeoffs were inevitable. Explaining games is a common, social practice, making it an accessible domain for a large variety of participants. The occasion to explain a game creates a setting for natural explanations and lessens any perceived pressure or exam-like atmosphere. Regarding participant acquisition, one can learn the game quickly, whereas mastering the game can be considered challenging. As \textit{Quarto!} provides strategical depth, it allows for some sort of variance in the explanations. Yet, the scope of what one may explain is narrow enough to make the explanations comparable. 

As games are invented by humans, they are technological artifacts. We believe, games exhibit their dual nature quite well in the context of explanations: One has to explain the pieces, board and rules, hence the Architecture. But one also needs to explain how these things go together and create an interesting game in which two players struggle to win. Comparable to chess, for example, one could explain how certain moves can be interpreted as offensive or defensive. In theoretical terms, the explanation also needs to elaborate how one follows their intentions within the game and thus address the game's Relevance. In summary, providing an explanation that results in a deep understanding of a game does not only require explaining its components and rules (knowing what, Architecture), but also explaining why certain aspects of the game are important to consider during gameplay (knowing why; Relevance). This is comparable to understanding XAI explanations.

The decision for the absence of the game had different reasons. Regarding the theoretical framework for the analysis, we expected having the game on hand would limit utterances that address the Architecture of the game (e.g., about the different shapes of the pieces). Thus, the EX needs to decide which detail requires explanation, allowing us to perceive strategies that EX use to, for example, generalize such details. As gestures are linked to or results of cognitive processes, a gesture analysis offers additional insights to get an even more detailed understanding of what happens during explanations at later stages of the project \cite{mcneill_hand_1992,mcneill_gesture_2008}. As a consequence, the table between the participants needs to be free of any objects to not influence their gestures (e.g., participants could hold on to pieces of the game).
\subsection{Procedure}
\label{sec:acquisition}
The study was conducted by researchers and research assistants from an interdisciplinary background (linguistics, psychology, and computing education). After participants arrived, they were welcomed, and, for the sociodemographic questionnaire, led into two different rooms independently of each other. During the study, all communication between researchers and participants was based on boilerplate, too. In phase 2, the explanations were recorded from three different camera angles: a view of the table, a medium-shot of the EE upper body, and a medium-shot of the EX upper body. 

After EE and EX filled out the questionnaires for sociodemographic details, they were instructed about the next phase. The EX's instruction was: "In the next room, please explain the game to the person across from you in such a way, that the person would have a chance to win the game"\footnote{Original: "In dem nächsten Raum erklären Sie bitte dem Gegenüber das Spiel so gut, dass Ihr Gegenüber eine Chance hätte, das Spiel zu gewinnen"}. The EE's instruction was, "[i]n a moment, a person will enter the room and explain a board game to you. Please participate actively in the explanation" before the EX entered the room\footnote{Original: "Im nächsten Raum ist eine Person, die Ihnen ein Spiel erklärt, bitte nehmen Sie aktiv an der Erklärung teil"}. 

In phase 2, the EX explains the board game \textit{Quarto!} to the EE sitting vis-à-vis at a table. During the explanation, the game is \textit{not} present due to reasons given in Section ~\ref{sec:quarto}. After the instruction, the EX entered the room of the EE, sat down across from the EE and a researcher gave a starting signal, left the room, and the explanation phase started.

For the video recall, we observed the explanation via the livestream and took notes of important scenes during the explanation. We used an identification scheme for the ad hoc selection of scenes to be used in the video recall. Criteria for selection were: substantial contributions, self- and other initiated repairs, turn taking, misunderstandings and questions that assessed understanding and knowledge of the EE. After the explanation ended (phase 2), we quickly started the video recall with the EX to investigate the EX's thoughts regarding specific points in the explanation \cite{fox-turnbull_autophotography_2011}, especially to discover reasons for addressing either of the two sides of the dual nature in their explanations. 

For the video recall with the EX, we used a semi-structured interview which included questions about explanation content, explaining intention, explanation quality, the assumed EE's understanding of the game, and perceived knowledge needs of the EE. During the video recall, a researcher watched the videotaped explanation with the EX from start to end, and stopped at the previously selected scenes. This served as a stimulus to enable the EX to give detailed insights into certain sequences of interaction, their interpretation of a particular moment \cite{fox-turnbull_autophotography_2011} by allowing participants to not solely rely on memory when trying to answer the questions\cite{lyle_stimulated_2003}. Questions were answered before information has been transferred from short- to long-term memory to avoid conflation of experiences and conjoined memories \cite{mackey_second_2005} by starting phase 3 as quickly as possible. The whole video call procedure was standardized. 

\subsection{Analyses}
\label{sec:analysis}
This paper puts the focus on the analysis of the explanation of the game, especially on what was said  and why with regard to the dual nature of the explanandum. An analysis of the explanations content provides the "what was said at a certain time". That analysis is, in a second step, supported by an analysis of the video recall to support the interpretation by investigating the underlying reasoning. The video recall data provides the EX justifications, and thus helps to better understand the development of the explanation regarding the reasons for utterances addressing Architecture and Relevance at certain times. 
\subsubsection{Explanation Content Analysis}
\label{sec:dialogue_analysis}
The explanations recorded in phase 2 served as the foundation for this analysis. Student assistants and colleagues from linguistics transcribed all recordings of the explanations according to the second complexity level (basic transcript) of the GAT2 transcription system \cite{selting_gesprachsanalytisches_2009}. The GAT2 transcription system level 2 segments the spoken text into intonational units. The length of the transcripts was between 118 and 653 intonational units (\textit{M}=332, \textit{SD}=146). In the final transcript, an intonational unit represented one line in the transcript.

We used qualitative content analysis \cite[p. 70]{kuckartz_qualitative_2014} to analyze the transcripts. The code system consisted of two deductive code categories: (1) utterances addressing the Architecture, and (2) utterances addressing the  Relevance of the explanandum. We developed a coding manual over the course of multiple pilot studies during which the first and second coder (first author and a student assistant) discussed their coded segments in data sessions, and refined the coding manual. All transcripts were double coded using the final version of the manual. All coded segments were additionally labeled to represent whether the utterance was by the EX or EE. Inter-rater and intra-rater reliability were calculated using MaxQDA, which reports the Brennan and Prediger Kappa \cite{brennan_coefficient_1981}. Across all transcripts, the inter-rater reliability between both coders was "Almost Perfect" (\textit{k}=0.80) \cite[p. 165]{landis_measurement_1977}. The intra-rater reliability was calculated for both coders who coded four transcripts twice three weeks apart (\textit{k}=0.91). The performance of the coding manual is satisfactory for the aim of the study. Overall, the content of the segments coded with the two code categories align with the theoretical ideas, and thus we were able to identify which sequences of the explanations contained utterances addressing the Architecture or Relevance reliably.
\begin{wraptable}{L}{7.5cm}
\caption{\textit{Overview of the Code Categories used for the Analysis of the Explanations' Content}}
\begin{tabularx}{\linewidth}{@{}p{.75in}X@{}}
\toprule
\textbf{Code} & \textbf{Examples} \\ \midrule
\textit{Utterances addressing Architecture ((UA)}  &   EX °h and they will then be ALTernately- \newline 
   put onto this BOARD, \newline(P01, Pos. 55-56) \newline  \newline     if I would STARt now \newline 
    i would have to hand YOU one [a piece]  \newline(VP22, Pos. 213-214)  \newline   \newline  
        and these PIECES have \newline 
    FOUR properties \newline 
    they are either uhm TALL " \newline  
    or SMALL \newline(VP06, Pos. 27-30)\\ \midrule
\textit{Utterances addressing Relevance (UR)}    &       EX  SO what makes this game interestING \newline is that uhm \newline  if YOU plAced one [piece] \newline then YOU decide \newline     which piece your opPOnent places next \newline(VP02, Pos. 204-208)\newline \newline  \\ \bottomrule
\end{tabularx}
\label{tab:code-category-content}
\end{wraptable}

A total of 878 segments were coded across all transcripts (605 utterances addressing the Architecture, and 273 utterances addressing the Relevance of \textit{Quarto!}). On average, 56\% (\textit{SD}=10,37\%) of the transcripts were coded, while 44\% (\textit{SD}= 10,12\%) received no code. This is mainly due to the transcription system which makes use of spaces and special characters to include additional information (e.g., pauses, breathing, overlaps). Another reason is that most explanations contained some small talk which was not part of the explanation. Out of 605 coded segments, 2,62\% received both codes, as either one or the other category was possible due to different plausible interpretations. Examples for that were transcripts starting with an example ("the GA:ME \// is a varIANT \// of coNNECT four" (VP11, Pos. 5-7)\footnote{All examples we use from the transcripts of this paper were carefully translated from the German original to English by the first and second author of this paper.}), or areas in which a shift from Architecture to Relevance occurred within one intonational phrase and thus both codes needed to be applied. In general, the content of each segment corresponded well (98\%) to one of the code categories, thus allowing us to consistently identify whenever the Architecture or Relevance were addressed during an explanation. For an overview of the contents of each code category, see Table ~\ref{tab:code-category-content}. 

For the analyses, we used MaxQDA and Python; MaxQDA for coding, reliability calculations and the majority of qualitative work, and Python in a second step, especially for visualizations and calculations, to investigate patterns regarding combination of sequences of the two code categories. For these visualizations, all transcripts were standardized in length. The starting point was always the first line in the transcript, that was coded. The end was the last line which received a code. For the standardization, all transcripts are divided into 100 sections with equal length and labeled with regard to the codes used in each of the 100 sections, similar to how MaxQDA realizes this feature\footnote{If, for example, one section contained three lines of transcripts, one addressing the Architecture and two addressing the Relevance, the visualization represented this by drawing a rectangle which is to 1/3 colored according to the code Architecture and 2/3 colored according to the code Relevance in order of their occurrence in the transcripts. In other words, there is no large loss of precision in the visualizations as a result of the standardization of length.}. We used these visualizations to identify specific points of interests in the corpora of explanations (e.g., switches from Architecture to Relevance, EE remarks regarding Relevance after a long phase of utterances about Architecture by the EX and vice versa). These points of interest served as a guide for the selection of video recalls we analyzed in the next section, which are used to underpin the analysis of the content of explanations. 

\subsubsection{Video Recall}
\label{sec:videorecall}
To investigate the justification for the content of the explanations as perceived by the EX, this analysis focuses on the interview questions from the video recall. The video recall consisted of a larger variety of questions. Thus, only questions regarding the EX's explaining intention, as well as the EX's perceived knowledge needs of the EE are analyzed. A student assistant transcribed the data from ten video recalls using standard orthography \cite{oconnell_transcription_2009}.  

\begin{wraptable}{R}{6cm}
\caption{Category System to code the reasoning for explanation content stated by EX during the video recall}
\label{tab:code-categories-vr}
\begin{tabular}{lc}

\toprule
\textbf{Code Category} & \textbf{Subcategory} \\ \midrule \midrule
\textbf{Architecture}  &                    \\ \midrule
              & Knowledge Needs of EE                  \\\midrule
              & Explaining Intention of EX         \\ \midrule\midrule
\textbf{Relevance}     &                           \\ \midrule
              & Knowledge Needs of EE               \\\midrule
              & Explaining Intention of EX             \\ \bottomrule
\end{tabular}
\end{wraptable}
We used a deductive category system to categorize reasons for explanation content in consideration of the dual nature theory, and to categorize the reasons into two subordinated code categories: Knowledge Needs of the EE, and Explaining Intention of EX (see Table ~\ref{tab:code-categories-vr}). Depending on whether a statement of the EX provided reasons for addressing "Architecture" or "Relevance" in their explanations, either the perceived knowledge needs of the EE or the explaining intention of the EX were coded with the corresponding sub-category. Statements which could not clearly be assigned to Architecture or Relevance were collected under a main category (Reasons for Explanation Content). The questions for the semi-structured interview guideline were derived from literature. We formulated the questions so that the explaining intention of the EX (why the EX wanted to explain specific aspects in certain situations irrespective of the situation, or the EE), as well as the knowledge needs of the EE as perceived by the EX, were focused on in the answers. We considered both of these aspects as important factors for the contents of explanations \cite{chi_can_2004,heider_psychology_1958,levelt_speakers_1981,roscoe_self-monitoring_2014}.  

We coded the material using a coding manual that we developed beforehand. The length of units of coding was flexible, ranging from at least one phrase to sometimes several sentences. Most important was that the unit encapsulated one specific reason stated by the EX (EX's explaining intention, or EE's knowledge needs). If multiple reasons were mentioned within a one answer to a question, each reason was assigned to a separate unit of coding \cite{schreier_qualitative_2012}. All video recall transcripts were coded twice, and intra-rater reliability was "substantial" (\textit{k}=0.72)\cite[p. 165]{landis_measurement_1977}, which is satisfactory as the data were complex. Only ten video recalls are part of this analysis, as the final solution for the video recall was not fully developed until the second half of data collection, especially due to finding a reliable technical solution.

\section{Results}
We start this section by presenting results from the analysis of the content of the explanations. Afterward, to underpin the commonalities we found across the explanations, we support the findings with the results from the analysis of the video recalls. 

\subsubsection{Explanation Content}
\label{sec:explanationcontent}
The investigation of RQ1 (How is the dual nature of the explanandum expressed in the utterances of the explanation?) based on a qualitative content analysis. As an analytical framework, the techno philosophical theory of the dual nature provided two deductive code categories, which we used for coding the corpus. 

\begin{figure}
    \centering
    \includegraphics[width=\linewidth]{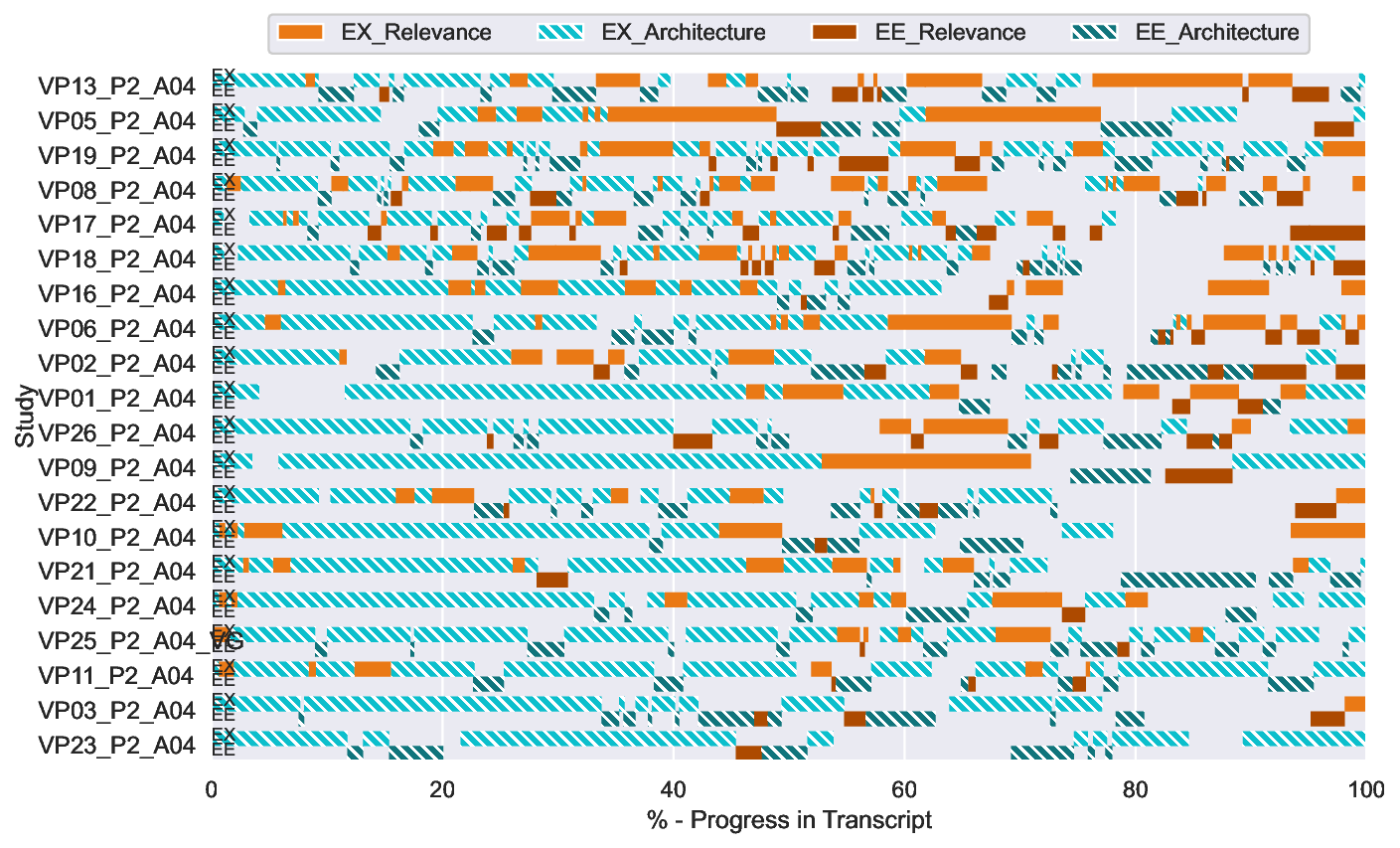}
    \caption{Visualization of explanation content across studies. All transcripts standardized in length. Start- and endpoint are the first and last coded segments in the transcripts. Sorted by utterances addressing Architecture (least to most; EX = Explainer, and EE = Explainee) }
    \label{fig:arch_sort}
\end{figure}
Except for the EX in VP23, all EE and EX addressed both sides of the dual nature in the corpus (see \figurename~\ref{fig:arch_sort}). In general, the content of the coded segments within each category aligned well with the ideas of the dual nature theory. When the Architecture of \textit{Quarto!} was addressed, EX and EE alike, used, for example, causal reasoning to address certain rules, described how the rules of the game result in a certain sequence in which each of the two players has to act, or described the physical makeup of different parts of the game. Whenever they addressed the aspects of Relevance of \textit{Quarto!}, they shifted to teleological aspects, for example, that the complexity rises continuously throughout the game ("that means the MORE pieces are put onto the board (0.7) \// the MORE you need to thInk \// because there are MULTiple rows containing three pieces", VP25, Pos. 174-176), strategical recommendations (" a:nd it is a: stra\_strategical GAME \// because if a fiew pieces are PLACed \// then YOU are able to see \// which PIECE \// would be GOOD \// to hand to me \// such that i would not WIN \// that means you also have to thiNK a little" VP16, Pos. 75-82), and emotional aspects for example how it is especially annoying to personally hand the piece to the opponent who may win using that piece in their next move ("EX (0.3) the frusTRATing bit about the game is that \// that ONESELF is always the reason \// for loSING \// because one HANDS OVER (.) a piece" VP18, Pos. 372-375). A summary of the contents of the two categories, as well as examples from the transcripts, is provided in Table \ref{tab:code-category-content}.

\begin{wrapfigure}[11]{r}{0.6\textwidth}
    \includegraphics[width=\linewidth]{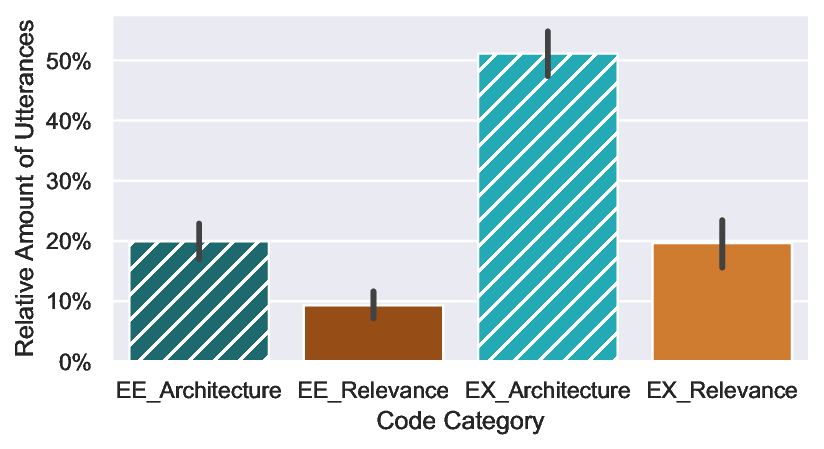}
    \caption{Average code distribution across studies}
    \label{fig:code_ratio}
\end{wrapfigure}

\textit{Architecture Addressed More Frequently:} Both EX and EE addressed each side of the dual nature in their explanations, but the balance of which side was focused on shifted throughout the explanations. While some EX (e.g., VP23, VP03) were focusing almost exclusively on aspects of the Architecture, other EX addressed aspects of the Relevance of \textit{Quarto!} more often (e.g., VP13, VP05). The amount of talk EX and EE spent addressing each of the sides varied quite a lot across dyads, but on average, the Architecture was addressed more frequently than the Relevance for EX and EE (see \figurename~\ref{fig:code_ratio}). 

\begin{wrapfigure}{l}{0.55\textwidth}
    \includegraphics[width=\linewidth]{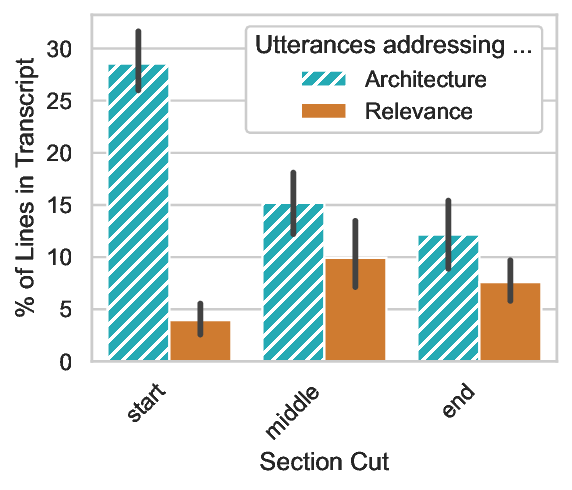}
    \caption{Comparison of the percentage of lines in the transcripts addressing the two code categories by speaker sectioned in thirds.}
    \label{fig:section_cut}
\end{wrapfigure}

\textit{Architecture First -- Laying out the Tools?}
In the context of RQ2, we were especially interested in whether we would find any patterns across explanations regarding \textit{when} and \textit{in which configurations} the sides of the dual nature were addressed. A comparative visualization of all transcripts of the corpus grouped by utterances by EE and EX and colored according to which side of the dual nature was focused on throughout the transcripts, see \figurename~\ref{fig:arch_sort}. As indicated by the visualization, utterances addressing the Architecture were uttered more frequently overall. Especially in the early stages of the explanations, the aspects of the Architecture were addressed more frequently than in the middle and last third (cf. \figurename~\ref{fig:section_cut}). 

The plot \figurename~\ref{fig:section_cut} provides additional evidence that  EX referred to aspects of the Architecture in the first third of their explanation more often than in other parts of the explanation. In the explanations, the content of the first third, especially EX, seemed to address important components and pieces of the game, which they often referred to later in their explanations. We describe that phenomena as \textit{Laying out the Tools}. In the later stages of the explanation, the utterances are more balanced in regard to the side they address. In the second third of the explanations, utterances addressing Relevance were more prominent than in the first or last third. But utterances about the Architecture were dominating the explanations across all studies. Interestingly, if the dyads decided that the EE should share their understanding -- that happened rather frequently at the end of the transcripts -- the EE's reiteration addressed mainly aspects of the Architecture, too.

\subsubsection{Video Recall}
For answering RQ3 (How do EX justify their choice of explanation content regarding which side of the dual nature was addressed?) we used qualitative content analysis, too. Based on the results from \ref{sec:explanationcontent}, we selected 20 scenes from the video recalls: ten scenes focusing on the beginning of the explanation, and ten scenes after the content shifted from Architecture to Relevance.

\textit{Reasons for Architecture First} Reasons for addressing Architecture at the beginning of an explanation were found in the EX's explaining intention, i.e., what are the needs of the EX regarding the explanation content, and to a lesser degree in the EX reaction to EE's knowledge needs. From a total 28 reasons stated by EX for explaining content addressed to Architecture, 26 were rooted in the explaining intention of the EX irrespective of the EE. No EX mentioned the need to address aspects of the Relevance at the beginning. In two  instances (VP10, VP13), the EE expressed knowledge needs at the beginning of the explanation. Yet interestingly, the EX did not justify the explanation content by wanting to address these knowledge needs.
\begin{figure}
    \includegraphics[width=\linewidth]{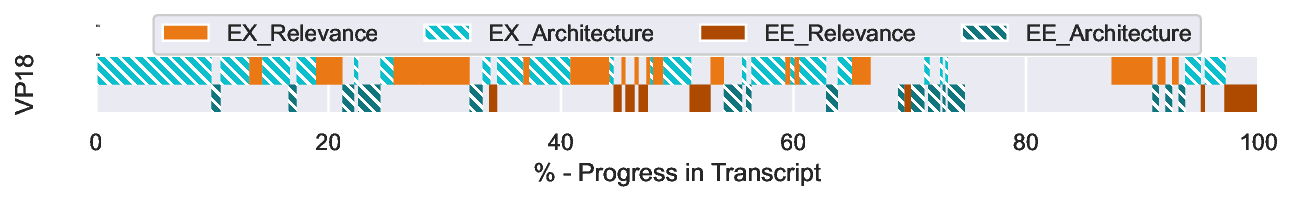}
    \caption{Combination of analyses of explanation content, and EX' justifcation for in the example of VP18. The EX addressed the Architecture in roughly the first 15\% and justified it using the following reasoning: "starting/ at first I thought it was important to know the components of the game before saying how (...) the game progresses" (VP18, VR1, Pos. 5)}
    \label{fig:vr_a_first}
\end{figure}

The reasons for focusing on the Architecture at the beginning (cf. \textit{Laying out the tools}) as stated by the EX were manifold. Most EX regarded it as important that the explanation has a certain structure, meaning that some aspects were believed to be prerequisites that need to be explained, before explaining further details. Two EX stated, for example, that explaining game components and their physical characteristics was a requirement to be able to explain the goal of the game (VP10, VP18). In VP06, the EX reported: "In the beginning I wanted to explain the course of the game in a swift manner before explaining in-depth" (VP06, VR1, Pos. 3). Another reason for explaining Architecture first was found in the EX's wish to create an image of the game that the EE can imagine, and therefore explained game components and their appearance first. With a similar intention, one EX made a comparison to chess and described the game board as a smaller chessboard (VP21). In VP 24 the EX stated: "I just wanted to create a picture of the game so that he knows [...] what kind of game it is." (VP24, VR1, Pos. 4). 

Even though the EX addressed primarily Architecture in the beginning and thus Relevance is not addressed in the explanations, we found indicators that Relevance might already be part of an explanation, even though it can only be determined in the EX justification for explanation content. In these cases, Relevance is not expressed verbally, yet aspects of the Relevance had an impact on the structure of the explanation (e.g., VP13, VP17). One EX explained in detail the different pieces and their similarities and justified it by addressing the Relevance in the interview by saying: "Because that's what the goal of the game actually means to me. To find these similarities." (VP17, VR1, Pos. 5), addressing the Relevance by implying that getting four in a row is \textit{not} important, but instead finding the similarities on the board is. 

In summary, reasons for addressing the Architecture in the beginning rooted predominantly in the EX's explaining intention and covered, for example, structuring the explanation, as well as pictographic descriptions of main aspects.

\textit{Reasons for shifting to Relevance after addressing the Architecture:} Reasons for shifting to explaining Relevance in later stages of the explanation were rooted in both the EX's explaining intention and the perceived knowledge needs of the EE. In total, 17 reasons in the 10 scenes included reasons for addressing the Relevance, whereof nine were related to knowledge needs of the EE and 8 were related to the explaining intention of the EX. 

\begin{figure}
    \includegraphics[width=\linewidth]{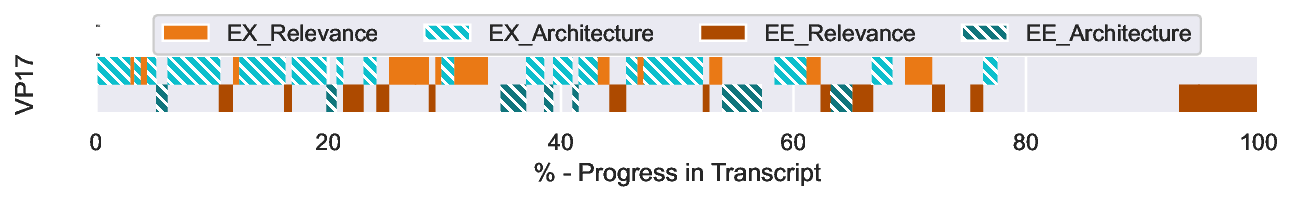}
    \caption{Combination of analyses of explanation content, and EX's justification in VP17. The EX shifted from Architecture to Relevance between 20\% and 40\% of the explanations progress, and justified it using the following reasoning: "My counterpart wanted to know why or what the rationale is for the fact that one decides for the opponent [which piece to play]. Who then could have won the game using that piece.  Which he would have handed to me [and thus I can win instead]." (VP17, VR3, Pos. 7)}
    \label{fig:vr_ar_shift}
\end{figure}

There were three main reasons for addressing the Relevance rooted in the knowledge needs of EE, ranging from the desire to be able to assess the game's degree of complexity (VP6, VP24), to certain aspects that are important to consider when trying to build a row of four (VP25), and the expressed wish to learn more about the game's strategies (VP13). The EX's explaining intention, resulted in explanation content that was addressed towards Relevance, too (e.g., VP10, VP18, VP26). The EX of VP26 referred to gameplay experience: "It was about the exchange of my experience, it is not about knowledge, but about experience and about situations experienced during a round or the game." (VP26, VR 5, Pos. 9). There were instances within the explanations for which the reasons could not be categorized to EE knowledge needs or EX explaining intention. In these cases, the reasons for addressing the Relevance in the explanation emerged from the interaction between both EX and EE. Sometimes, for example, questions were posed by the EE that addressed the Architecture and the EX replied addressing the Architecture but integrating information regarding the Relevance, too (VP21, VP22). In VP21, the EE asked what the end of the game looks like and the EX answered the question by referring to the Architecture, but also addressed to Relevance by mentioning the difficulty of keeping an eye on every game detail as the game progresses (VP21). Another example of how shifts to Relevance occur within an explanation is found in VP17, where the EX was explaining content addressed towards Architecture, specifically that players have to choose pieces for the opponent. The EE struggled to find the rationale for this, especially as it would mean that one would have to hand the winning piece to the opponent, too, and thus forced a shift to Relevance (see \figurename ~\ref{fig:vr_ar_shift}). 

The reasons for explanation content after shifting from Architecture to Relevance can be highlighted as highly diverse and for now, it can be said that reasons are rooted in EE's knowledge needs, the EX' explaining intention, and in questions emerging from the interaction between both EX and EE during the explanation.


\section{Discussion}
This paper investigated how the content of a naturalistic explanation of the board game \textit{Quarto!} evolves, and how this evolution of content is justified by the EX. Regarding XAI, we believe that researching how explanations of technological artifacts evolve naturally is an important prerequisite for formulating recommendations to construct synthetic explanations. One of the aims of this study was to investigate how the dual nature of the explanandum is addressed in the utterances of the explanations (RQ1). Very little was found in the literature regarding this question. The objective of the content analysis was to categorize utterances reliably into the two code categories: Architecture and Relevance. Throughout all explanations, both sides of the dual nature were addressed to a varying degree. Answering RQ2 (what are common patterns regarding sequences of utterances about the dual nature across the explanations?) resulted in a deeper understanding of when which side was addressed more frequently. We found that in most explanations, aspects of the game's Architecture were addressed in the first third of the explanation, and only in the middle and last third, aspects of the Relevance were addressed more frequently. Based on these findings, RQ3 (How do EX justify their choice of explanation content regarding which side of the dual nature was addressed?) helped to gain better insights to the reasons why EX explained predominantly the Architecture at the beginning of explanations, and how shifts from Architecture to Relevance were justified by the EX. We found that at the beginning of an explanation, especially the explaining intention of the EX, is the reason for explaining content addressed towards Architecture. In later stages, shifts from addressing Architecture to addressing Relevance were justified by the EX's explaining intention, the perceived knowledge needs of the EE, and their interactive sequences of EX and EE throughout the explanation. 

The study design worked as anticipated. All dyads engaged quickly into the explanation. The EX explained intuitively, and EE frequently asked questions or shared confusion. Our observation is that all dyads were interested in a successful explanation, and motivated to continue the explanation, until they thought to be finished. The comparability across explanations was satisfactory, as the time and content did not deviate too much and dyads kept their focus on explaining in such a way, that the EE would be able to win the game, too. Thus, instruction worked as well. Our overarching motivation is that finding certain patterns would provide insights useful for future research on other explananda to, eventually decide, for example, whether certain patterns are generalizable and maybe even a useful schematic for synthetic explanations, too.

As stated earlier, we believe, the theory provides a rich background for research on everyday explanations of technological artifacts. Based on our findings, we still believe that. It could provide a foundation for formulating recommendations on whether and which of the artifacts properties should be referred to within a (synthetic) explanation. In our study, EX tended to put more effort into explaining aspects that address the Architecture than the Relevance. In the case of \textit{Quarto!}, the questions of which side of the dual nature is more important or whether one is a precursor for the other (cf. Section \ref{sec:background}) can thus be answered: Explaining the Architecture first before diving into what EX considered to be more complex details of the Relevance of the game, seemed to be a natural way to approach an explanation in the context of \textit{Quarto!}. Regarding what we saw in the explanations, we would interpret it as a verbal realization of \textit{laying out the tools}, by addressing the bits of Architecture which the EX can refer to in later stages of the explanation to connect aspects of the Architecture of \textit{Quarto!} to aspects of its Relevance. Thus, we believe that, especially looking at the fairly large percentage of utterances addressing the Relevance in combination with the justifications that the EX stated in the video recalls, provide a solid foundation for the claim that \textbf{both} sides can be considered important in the case of \textit{Quarto!}. But whether and how these findings are generalizable, especially for more complex artifacts, is still an open research question. 

The question of how do EX (RQ3) justify their choice of explanation content regarding which side of the dual nature was addressed, brought further insights into the reasons why the EX addressed either the Architecture or Relevance. The EX provided a broad range of reasons for explaining aspects addressed to Architecture and Relevance at different stages of the explanation. The variety of different reasons indicate, that deciding on which aspects to explain is a process in which the EX has to consider a large variety of factors, not only concerning their intention but also guided by what they intend to explain. At the beginning of the explanations, EX dominated the explanation and the reasons for addressing the Architecture are usually rooted in the  EX's explaining intention. Therefore, the first part of the explanation follows the explanation goal and plan of the EX without further consideration of the specific needs of the EE, albeit important for understanding \cite{buhl_partner_2001,levelt_speakers_1981}. EX mentioned reasons like explaining physical components and their characteristics and appearance first as a preparation for providing more details later in the explanation, especially in regard to shifting to Relevance. At times, the shift from Architecture to Relevance did not occur suddenly, but emerged gradually in the context of the interaction of both participants. In these instances, even though the explaining intention of the EX was still apparent, the knowledge needs of EE became more evident. Reasons for that were not mentioned, but could potentially be out of general interest or due to identifying certain misunderstandings \cite{chi_can_2004}. Aspects that were addressed to Relevance were seldom part of the first third of the explanations. Yet, our analyses of the video recall showed that in some instances, EX structured their explanations with aspects of the Relevance in mind. They interpreted these bits of information as highly important for their explanation. This might be an indicator that Relevance is (1) an important part of an explanation and (2) even though verbally the Relevance is not yet addressed in the explanation, it still guides the choice of which aspects of  the Architecture are addressed at the start.  

The theoretical background puts focus on the characteristics of technological artifacts, but empirical investigations only addressed more technical, text-based explanations (e.g., patents) \cite{de_ridder_mechanistic_2006}. As texts are very different from extemporaneous explaining, we were surprised how clearly, and easily distinguishable Architecture and Relevance were addressed within the explanations of \textit{Quarto!}. In the cognitive sciences, the theory of people taking different stances to explain things is interesting in this regard \cite{dennett_intentional_1987,keil_birth_1994}. Two stances seem to be connected to addressing the sides of the dual nature. Whereas in the context of the dual nature theory, the question was raised whether one side is a precursor or more important, the cognitive sciences spent time researching how explanations from different stances result in different understanding.   If (and only if!) these stances would be connected to addressing either the Architecture or the Relevance, our findings \textit{may} support the hypothesis, that by categorizing utterances into either of the two code categories of the dual nature, in the case of \textit{Quarto!}, we identified areas in which these stances were observable. If this was the case, interestingly, our dyads shifted between stances fairly quickly and multiple times throughout one explanation. 



\subsubsection{Quality Standards}
We adhered to a variety of recommendations \cite{kuckartz_qualitative_2014,miles_qualitative_2014}. We documented the different stages of codings from initial codings until the final codings of the transcripts, especially for being able to reconstruct the interrater data sessions in which the codings were discussed between the two coders and adjusted accordingly. Inter-rater and intra-rater reliability was examined for the analysis of the explanation content, intra-rater reliability was assessed for the video recalls. Additionally, three independent coders unfamiliar with the theoretical background used the coding manual for the explanation content analysis  and coded three transcripts. We compared the results to our codings using inter-coder reliability tests (\textit{k}=0.53). All reliability measures were good (see Section ~\ref{sec:dialogue_analysis}). Parts of the final codings were discussed within and outside the research project (peer debriefing). Earlier results were presented and discussed during a research retreat of the TRR318. To prevent false conclusions, and for a deeper understanding of the practical implications of the study design, all researchers involved in this publication were present during at least a quarter of the studies (experiencing the study). We triangulated the results from the content analysis with EX justifications stated in the video recalls to underpin, contrast or contradict our interpretations. 



\subsection{Limitations and Future Work}
We found two commonalities across the explanations we analyzed. First, the majority of time was spent addressing the Architecture, especially in the first third of the explanation. Only later, the content was more balanced between utterances addressing the Architecture and Relevance. Whether this applies for other technological artifacts, too, is still an open research question that this study cannot answer. We believe that, especially if artifacts were more complex, describing the most important aspects of the Architecture at the beginning would maybe not be possible as (1) the amount of information could simply not be remembered (i.e., too much information) and (2) the amount of information would require explanations that would be far too long to be interesting for XAI. But, our insights open up new research opportunities and questions:
\begin{itemize}
    \item Is \textit{laying out the tools} a generally reasonable strategy when explaining technological artifacts? Do circumstances exist, in which it would make more sense to start with addressing the Relevance?
    \item How much focus on the Relevance is needed (and when) to provide a \textit{good} explanation?
    \item What explanation strategies are used to start the explanation, if the explanandum is a very complicated, technological artifact (i.e., what are other strategies opposed to \textit{laying out the tools})?
    \item How is the level of understanding of the EE related to certain patterns or the ratio between the utterances addressing the two sides in the explanations?
    \item How is the EE ability to competently play the game connected to knowledge about both sides conveyed in the content of the explanations?
\end{itemize} 

Based on our results, we now are in a position in which we claim that using the dual nature theory as an analytical tool to structure the content of explanations of technical artifacts is a worthwhile, and a viable approach. Coding the content of the explanations is quick, and the structure provided by the codes allows quick insights into how the explanation was carried out. Based on this, further analyses can be carried out which ultimately may help for recommendations for the construction of synthetic explanations, too. In the theoretical background elaborated in Section ~\ref{sec:background}, we addressed how different researchers have different perceptions regarding this question, yet more empirical support for the claims is needed. Our ongoing research will address such questions of which explanations, regarding their structure of addressing the dual nature of \textit{Quarto!}, are better, that is, result in a deeper understanding of the EE. Currently, we are in the process of developing an instrument for such an assessment. Supplementary, we will analyze phase 4 of the larger study, in which EX and EE play the game, which hopefully provides further insights regarding the level of understanding of the EE. 

One remark that is necessary is, that in the case of games, the need for knowing the Architecture of the game is inherently clear. For example, if one does not know the shapes of the pieces in \textit{Quarto!}, one would not be able to play it competently. Especially in the case of digital artifacts, this is not necessarily needed\footnote{Research areas such as user experience and interface design are a good example in this regard. Put very briefly and naively, they are interested in circumventing the need to understand the Architecture by providing enough Relevance to be able to use things competently}. Therefore, if researching explanations of digital artifacts, another layer of complexity is added, which most likely also influences how easy it is to distinguish which side of the dual nature is addressed. Yet, as a first step, by investigating an example in which the need to understand the Architecture is so prominent, we gained insights -- and still strive to gain more -- that could be fundamental for a deeper understanding of explanations of more complex, and especially digital artifacts.

To further support, contrast or contradict our results, our study design could be changed in a way in which either the EE or the EX is previously trained to influence the explanation by drawing the attention to either the Architecture or the Relevance, for example by instructing them to have certain knowledge needs (e.g., you want to program/build the game afterward vs. you want to know whether the game is a suitable present for a friend). Controlling that variable could be very promising to get an understanding if one side may also be omitted completely or if it would always emerge naturally, regardless of what was instructed before.

The video recall method has some limitations, for example, the anxiousness of the participants, self-censorship and reduced visual hints through fixed perspective and technical format \cite{calderhead_stimulated_1981}. We addressed this by creating an atmosphere in the lab that reduced factors leading to stress or anxiety in the participants and gave room for self-exploration. By using this introspective method, which offers numerous advantages, we were able to better identify the reasons for explaining specific aspects of the game. 

We had occasional statements in the interviews in which the EX were surprised by the absence of the game during phase 2. Therefore, the absence of the game is a potential factor influencing the beginning of the explanations quite drastically. But, as elaborated in \ref{sec:quarto}, we believe this circumstance to be the norm rather than the exception, as it created a black box scenario. Yet, this factor needs to be addressed in future research. In most explanations, a surprise regarding the absence of the game was not mentioned by the EX. Therefore, concerning the naturalness of the explanation, we believe it to have only a minor effect.

\subsection{Conclusion}
In the case of natural explanations of the board game \textit{Quarto!}, EX referred to a carefully selected set of information about the Architecture of the game in the first third of the explanation. Their rational was that such information (the \textit{What}) is a prerequisite for information that they considered to be more complex (the \textit{why}). They did this to make the more complex content of the explanation more accessible for the EE later in their explanations. At later stages of the explanations, the reasons for shifting from addressing the Architecture (the \textit{what}) to addressing the Relevance (the \textit{why}) were rooted in a larger variety of reasons, one of them being that they slowly emerged from the interaction between EX and EE. Our findings are a first step towards a more general understanding of extemporaneous explanations of technological artifacts, especially regarding when and how the two sides of the dual nature are addressed in explanations. A better conceptualization of these naturalistic explanations of technological artifacts can provide crucial insights into how XAI can explain its decision more naturally. Especially if certain patterns -- such as adding why to what -- are generalizable and thus could be embedded in an interactive explanation system, this could improve the interpretability of decisions, and ultimately support the agency of a large variety of people interacting with XAI.

\subsection*{Acknowledgements}
We thank our colleagues for feedback on the first drafts, as well as all research assistants for their support during coding, analyses, and data acquisition.

\subsection*{Ethics Statement}
This study was approved by the Paderborn University Ethics Board. All participants participated voluntarily and provided written informed consent prior to the studies. The studies were conducted in concordance with local COVID-19 policies.
\newpage
%
%
%

\bibliographystyle{splncs04}
\bibliography{literature}

\end{document}